\documentclass[letterpaper, 10 pt, conference]{ieeeconf}

\IEEEoverridecommandlockouts
\usepackage{cite}
\usepackage{amsmath,amssymb,amsfonts}
\usepackage{graphicx}
\usepackage{algorithm}
\usepackage{algpseudocode}
\usepackage{xcolor}

\definecolor{projectblue}{RGB}{0,82,155}
\usepackage{svg}
\usepackage{colortbl}
\usepackage{caption}
\usepackage{bm}
\usepackage{gensymb}
\usepackage{booktabs}
\usepackage{ragged2e}
\usepackage{hyperref}
\usepackage{cleveref}

\hypersetup{
    hidelinks
}

\usepackage{tabularx}
\usepackage{dblfloatfix}
\usepackage{placeins}
\usepackage{url}
\usepackage{etoolbox}
\usepackage{multirow}
\usepackage{multicol}
\let\labelindent\relax 
\usepackage{enumitem}
\usepackage{float}

\newcommand{\degs}{^{\circ}}

\makeatletter
\g@addto@macro\@maketitle{%
  \par

  \captionsetup{type=figure}
  \setcounter{figure}{0}
  \centering
  \includegraphics[width=2\columnwidth]{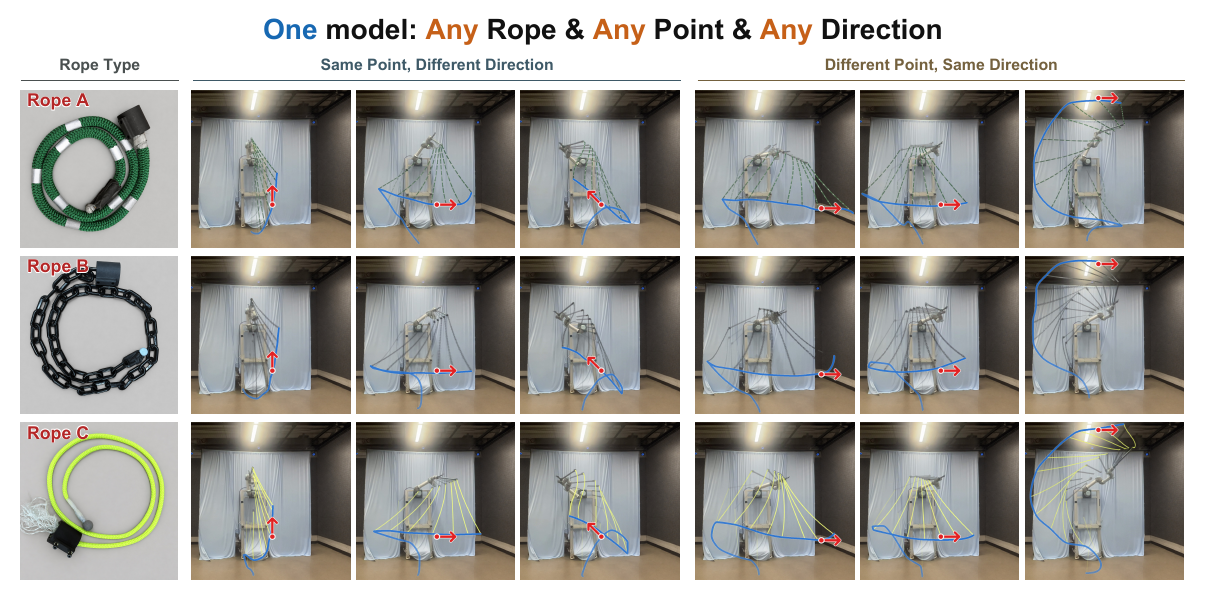}
  \captionof{figure}{
\textbf{One model, any rope, any point, any direction.}
Real-world strikes across three DLOs with different physical properties:
\textbf{Rope A}, a green braided rope;
\textbf{Rope B}, a black plastic chain; and
\textbf{Rope C}, a yellow braided cord.
For each rope, the robot strikes the same point from different target arrival directions (left) and different points with the same target arrival direction (right).
Red dots and arrows indicate the commanded strike position and arrival direction.
Each example is executed with a zero-shot strike.
  }
  \label{fig:figure1}
  \vspace{-5pt}
}
\makeatother

\algrenewcommand\algorithmicindent{1.0em}

\title{
\Large\bfseries
Point It, Strike It: Direction-Conditioned Dynamic Manipulation of \\Deformable Linear Objects\\
{\small\normalfont\sffamily\bfseries
\href{https://deformx.github.io/DeformY}{
  \textcolor{projectblue}{Project Page: deformx.github.io/DeformY}
}}
\vspace{-1em}
}
\author{
Yi Yang$^{1,2,3,\dagger}$,
Xiang Fei$^{1,\dagger, \ddagger}$,
Lehong Wang$^{1,\dagger, \ddagger}$,
Zilin Dai$^{4,*}$,
Ruogu Li$^{1,*}$,
Jiting Cai$^{1,*}$,
\\
Liyao Chang$^{1}$,
Xinyi Yang$^{1}$,
Henry Kou$^{1}$,
Ruijie Fu$^{1}$,
Lu Li$^{1}$,
Howie Choset$^{1}$
\thanks{$^{\dagger}$ Equal contribution; each reserves the right to be listed first.}
\thanks{$^{\ddagger}$ These authors jointly led the project.}
\thanks{$^{*}$ These authors contributed equally to this project.}
\thanks{$^{1}$ The Robotics Institute, Carnegie Mellon University,
5000 Forbes Ave, Pittsburgh, PA 15213 USA.}
\thanks{$^{2}$ School of Ocean and Civil Engineering,
Shanghai Jiao Tong University,
800 Dongchuan Rd, Minhang District, Shanghai, China.}
\thanks{$^{3}$ Zhiyuan College,
Shanghai Jiao Tong University,
800 Dongchuan Rd, Minhang District, Shanghai, China.}
\thanks{$^{4}$ John A. Paulson School of Engineering and Applied Sciences,
Harvard University,
29 Oxford Street, Cambridge, MA 02138 USA.}
}

\begin{document}

\maketitle

\thispagestyle{empty}
\pagestyle{empty}

\begin{abstract}

Goal-conditioned dynamic manipulation of deformable linear objects has
mainly specified goals as positions for a rope tip to reach. Many tasks,
however, depend on how the tip arrives. We therefore study single-swing rope striking with goals that specify the tip's 3D position and arrival direction, across the workspace and on
different ropes. This is challenging because rope dynamics are hard to model, no demonstrations exist, distinct
swings reach the same goal with different reliability, and the sim-to-real
gap extends beyond the rope. To address these challenges, we extend the
state-of-the-art DLO simulator DeformX with GPU acceleration, a stable
Cosserat rod solver, and a cross-flow aerodynamic model, yielding
DeformX~2.0, which is more than $20{,}000\times$ faster. We then propose
TRACE (Trace-rooted Adaptive Cross-Entropy), which generates striking data by warm-starting each new target from
the stored swing whose tip path passes closest to it. Its cost penalizes
rope bending and abrupt tip motion to favor repeatable swings. A conditional
flow-matching policy trained on this data reaches $92.1\%$ accuracy
in simulation. Finally, we propose RECAP (Residual Calibration Policy), which fits the simulator's rope and rig parameters to a few calibration swings and adapts actions with a correction policy trained in simulation. On a real robot, across three ropes, RECAP raises success within $5$\,cm from
$72\%$ to $87\%$ for position goals, and within $10$\,cm and $10^{\circ}$ from $50\%$ to $79\%$ for goals that also specify the arrival direction.

\end{abstract}

\section{INTRODUCTION}

Deformable linear objects (DLOs), such as ropes, cables, wires, and hoses,
are widely used in robotic and industrial settings~\cite{intro_survey,
deformx}. A line of work studies their \emph{goal-conditioned dynamic
manipulation}, where a robot swings or casts the object so that it reaches a
commanded goal~\cite{irp,lostarc,whip,wigglego,deformx}. The goals
considered so far mainly describe the object's position, typically a 2D or
3D point for the rope tip to reach. We extend rope-tip targeting to a 4D goal that specifies both the tip's 3D position and its arrival direction. Many tasks depend on how the tip arrives. For example, striking a spinning top in the direction of rotation helps maintain its spin, while striking against it can slow it down. Similarly, knocking an obstruction aside requires a strike that pushes it out of the way.

We consider a robot arm that holds one end of a rope. The robot receives a goal $g = (p^{*}, d^{*}) \in \mathbb{R}^{3} \times S^1$
and must execute a single swing whose tip passes through $p^{*}$
while moving along $d^{*}$. The task covers any point in the reachable
workspace, any arrival direction, and any rope. Even for a person, hitting a point with the tip of a rope requires deliberate practice, and reaching it from a specified direction is harder still. To the best of our knowledge, no prior rope-hitting method addresses a task of this difficulty. For robots, this task is challenging for several reasons:

\begin{itemize}[leftmargin=*,labelsep=0.4em]
\item \textbf{Complex dynamics and object properties:} a swung rope is a
stiff, nonlinear, aerodynamically damped continuum whose tip amplifies every
command error, and its response depends on properties that vary between
ropes.
\item \textbf{Data sparsity:} no demonstration corpus exists for this task,
and it is hard enough for people that teleoperation is impractical.
\item \textbf{Many-to-one solutions:} very different swings can hit the same
target, differing in wind-up, swing plane, or timing.
These swings are not equally reliable either.
Some hold their tip path under small disturbances, while others change
drastically when replayed. 
\item \textbf{Sim-to-real gap:} a swing that works in simulation misses on
the real robot, and the mismatch is not confined to the rope. Controller lag, mechanical vibration, and timing error also contribute to this mismatch.
\end{itemize}

To address these challenges, we first extend DeformX~\cite{deformx} with GPU acceleration, a stable Cosserat rod solver~\cite{cosserat}, and a quadratic cross-flow aerodynamic model~\cite{air_drag,flycast}, yielding an accurate and efficient DLO simulator we call DeformX~2.0. We then propose
\textbf{TRACE} (Trace-rooted Adaptive Cross-Entropy) to generate striking
data entirely in this simulator. Starting from one swing verified on
hardware, TRACE warm-starts the cross-entropy method for each new target
from the stored swing whose tip path passes closest to it. Its cost favors
an extended rope and smooth tip motion, so the swings it keeps stay
repeatable. A conditional flow-matching policy~\cite{fm} then learns to
strike from this data alone. The network learns to generate suitable swing patterns for different goals, with goal conditioning in every network block to improve striking accuracy~\cite{dit,film}. We refer to its output as the \emph{nominal action}. Finally, \textbf{RECAP} (Residual Calibration Policy) adapts nominal actions to a real rope using a few calibration swings to identify the rope and rig parameters. These estimates condition a simulation-trained correction policy, and the fitted simulator selects the best candidate from the nominal action and its corrected variants. Calibration is performed once per rope, after which each goal requires a single zero-shot strike. Our main contributions are:

\begin{enumerate}[leftmargin=*,labelsep=0.4em]
\item \textbf{DeformX~2.0}, which extends DeformX~\cite{deformx} with GPU
acceleration, a stable Cosserat rod solver, and a quadratic cross-flow
aerodynamic model. It simulates ropes more than $20{,}000\times$ faster
than DeformX.
\item \textbf{TRACE}, a data generation method that produces smooth striking data entirely in simulation. In simulation, TRACE covers $98.3\%$ of the workspace, against $49.5\%$ for CEM with an example swing.
\item \textbf{4D goal-conditioned flow-matching policy}, to our knowledge the first rope-striking policy conditioned on 4D goal. It injects the goal into every residual block through zero-initialized modulation, which keeps the distinct swings that reach a goal apart instead of mixing them. Learned from simulation data alone, it reaches $92.1\%$ accuracy within $5$\,cm and $10\degs$.
\item \textbf{RECAP}, which combines system identification with a simulation-trained correction policy to adapt nominal actions to a real rope. On hardware, across three ropes, it raises success within $5$\,cm from $72\%$ to $87\%$ for position goals and from $28\%$ to $52\%$ for position-and-direction goals.
\end{enumerate}


\begin{figure*}[t!]
    \centering
    \includegraphics[width=\textwidth]{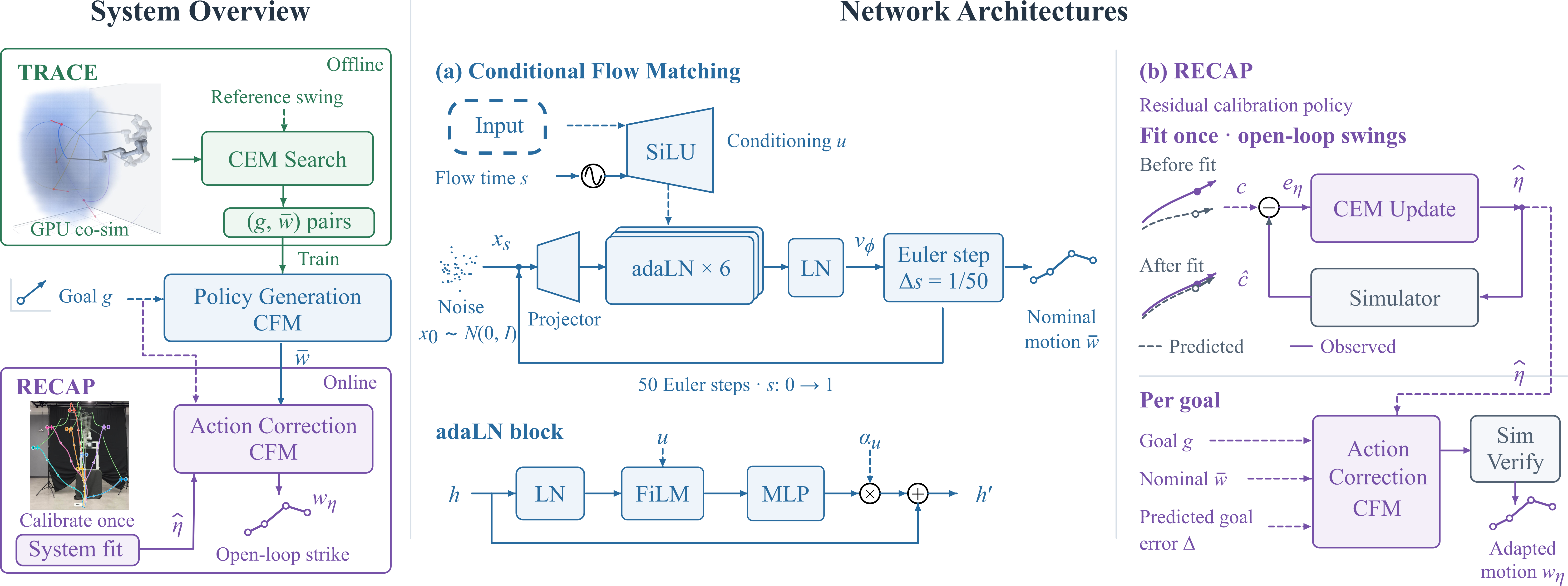}
    \caption{\textbf{System overview and network architectures.} Left: TRACE generates goal--action pairs in GPU co-simulation to train the base policy. RECAP calibrates the real system once and corrects the nominal action for each goal before a single open-loop strike. (a) The base and correction policies use conditional flow matching (CFM), conditioned on $g$ and $(g, \bar{w}, \Delta, \hat{\eta})$, respectively. (b) RECAP iteratively fits rope and rig parameters to measured calibration swings. These estimates condition the correction policy to adapt actions to the real system. Dashed lines indicate input.}
    \label{fig:network_architectures}
\end{figure*}

\section{RELATED WORK}

\subsection{Goal-conditioned dynamic rope manipulation}

Robots have learned to swing ropes and cables toward commanded goals.
IRP~\cite{irp} hits 2D targets with a robot-held rope by correcting the action
over repeated real-world attempts. Whip primitives~\cite{whip} reach 3D points
with low-dimensional open-loop motions, and Lost Arc~\cite{lostarc} vaults,
knocks, and weaves a cable around obstacles. Wiggle\&Go~\cite{wigglego}
strikes 3D points on new ropes. It predicts the rope's simulation parameters
from a short wiggle with a network trained in simulation, then optimizes the
motion for each goal in the simulator. DeformX~\cite{deformx} reaches 2D
targets in a fixed plane with per-step reinforcement learning. These goals mainly describe where the tip
should be and leave how it arrives unconstrained. Several of these systems are
also restricted to a plane or need repeated real attempts for each goal. We
specify the goal by both the 3D position of the tip and its arrival
direction, and reach any such goal in the workspace with a single zero-shot strike.


\subsection{Planner-generated training data}

When demonstrations are scarce, planners can generate training data instead.
Zhu et al.~\cite{zhuplanners} show that sampling-based planners produce
inconsistent demonstrations on which even expressive diffusion policies
degrade. PhysicsGen~\cite{physgen} obtains consistent data by re-solving
trajectory optimization warm-started from a few demonstrations under perturbed
conditions. Dynamic rope striking adds two difficulties. There is no
demonstration set to warm-start from, and swings that reach the same goal
differ in how reliably they transfer to hardware. TRACE starts from a single swing verified on hardware and warm-starts each new target from the stored swing whose tip path passes closest to it, which keeps the solutions for nearby goals consistent. Its cost favors an extended rope and smooth tip motion to keep swings repeatable, and each verified swing is labeled with its achieved outcome as in hindsight relabeling~\cite{her,gcsl}.

\subsection{Generative policies for multimodal action distributions}

Diffusion Policy~\cite{diffpolicy} and flow matching~\cite{fm,icfm} model
distributions over actions instead of a single prediction.
DA-MMP~\cite{dammp} builds on motion manifold primitives~\cite{mmfp} and
applies conditional flow matching to goal-conditioned throwing. In rope
striking, distinct motion patterns reach the same goal, making the action distribution inherently multimodal.
A regressor tends to average across these patterns, often producing an invalid swing. A generative model avoids this average, but its samples still spread between patterns when the goal enters only at the network input.
Our flow-matching policy injects the goal into every residual block through zero-initialized modulation~\cite{dit,film}, which keeps its samples within a pattern and raises one-shot accuracy.

\subsection{Sim-to-real calibration}

One line of work identifies physical parameters and uses them downstream.
Wiggle\&Go~\cite{wigglego} predicts rope parameters with a learned network and
optimizes each goal in simulation with them, and
GenDOM~\cite{gendom} conditions a policy on parameters estimated from one real
demonstration. Another line corrects actions from observed outcomes.
TossingBot~\cite{tossingbot} learns a residual on a ballistic release
velocity, and IRP~\cite{irp} refines the swing action over repeated real
attempts at each target. Our proposed RECAP fits rope and rig parameters to a few calibration swings for each rope, and a delta policy trained in simulation and conditioned on these parameters corrects the base action for the error the fit still misses.


\section{Problem Formulation}

A robot holds one end of a rope and executes a single swing.
Given a target position and arrival direction, the goal is to generate a
robot trajectory whose rope tip reaches the target while moving in the
desired direction.

The input is a goal:
\begin{equation}
    g = (p^*, d^*) \in \mathbb{R}^3 \times S^1,
    \qquad
    d^* = (\cos\theta^*, \sin\theta^*).
\end{equation}
where $p^*$ is the desired tip position, $d^*$ is the desired
arrival direction, and $\theta^*$ is the desired arrival angle. We measure arrival angle after projecting the tip velocity onto the tangent plane of the arm-centered sphere at the target. The up vector represents the corresponding unit direction in that plane, as shown in Fig. \ref{fig:coverage} (b). The goal thus has three position coordinates and one angular degree of freedom. 

The output is a parameter vector $w \in \mathbb{R}^m$ that decodes into an executable joint trajectory
$q_w(t)$ over a fixed time horizon. The action $w$ specifies joint-space
waypoints and a duration. A clamped cubic spline through the waypoints
reconstructs $q_w(t)$, and the duration is adjusted so that the trajectory
satisfies the joint-velocity limit.
The rope’s response to a robot action depends on the properties of the rope and robot setup, which we denote collectively by \(\eta\).

A goal does not uniquely determine a swing. Distinct robot trajectories may
bring the rope tip to the same goal.
We therefore represent the possible swings with a conditional distribution
$p_\eta(w \mid g)$ over feasible trajectories.




\section{METHODS}

\subsection{Overview}

We decompose the problem into nominal inverse learning and rope-specific adaptation. First, we use a simulator with nominal rope properties $\bar\eta$ to generate feasible goal--trajectory pairs. A conditional flow-matching policy
\cite{fm} then learns the nominal inverse distribution
\begin{equation}
    q_\phi(w \mid g)
    \approx
    p_{\bar\eta}(w \mid g).
\end{equation}
This stage captures the general relationship between desired rope-tip
outcomes and robot trajectories.

Second, the sim-to-real stage uses a few calibration swings to characterize the real rope system $\eta$. Action adaptation $T_\psi$ then corrects a nominal action $\bar{w}$ with the goal $g$ and the calibrated system:
\begin{equation}
    w_\eta = T_\psi(\bar{w}, g, \eta).
\end{equation}
Calibration is performed once per system, after which each goal is
executed with an open-loop swing.



\subsection{Learning robust motion patterns}
\label{sec:sim_learn}

Dynamic rope manipulation is sensitive to small changes in robot motion, yet distinct swings can reach the same goal. These swings may differ in their wind-up, swing plane, or timing. We refer to groups of similar swings as \emph{motion patterns}. Within some patterns, changes in robot motion produce gradual changes in the rope trajectory. Interpolating between different patterns, however, can produce unsuccessful swings.

This multiplicity motivates a generative policy rather than deterministic regression. A regressor trained with squared error tends toward the conditional mean, which may average incompatible swings. Conditional flow matching~\cite{fm} instead models a distribution of actions for each goal, allowing it to represent distinct motion patterns.

\subsubsection{\textbf{Data generation}}
\label{sec:data_gen}

We call our generator \textbf{TRACE} (Trace-rooted Adaptive Cross-Entropy). It uses the cross-entropy method (CEM)~\cite{cem} to generate accurate strikes while favoring robust rope motion. We guide CEM with:
\begin{equation}
\label{eq:cost}
J(w)=L_{\mathrm{target}}+L_{\mathrm{spd}}+L_{\mathrm{lim}}+L_{\mathrm{curv}}.
\end{equation}



The target, speed, and limit terms penalize position and direction errors, insufficient tip speed, and constraint violations. We accept a swing when these normalized losses satisfy $M(w)=\max\{L_{\mathrm{target}},L_{\mathrm{spd}},L_{\mathrm{lim}}\}\le1$. The curvature term \(L_{\mathrm{curv}}\) favors an extended rope and smooth tip motion, encouraging repeatable swings that tolerate disturbances and sim-to-real mismatch.

Our proposed TRACE keeps the solutions for nearby goals consistent by choosing where each search starts. We initialize the search with a manually tuned swing, which has verified robustness on hardware, providing guidance for motion properties not fully captured by the objective. Each new target is warm-started from the stored swing whose tip path passes closest to it, rather than the swing that solves the nearest target. This encourages CEM to extend an existing motion pattern across nearby goals. A final refinement pass re-solves targets from neighboring actions and keeps only verified solutions. Additional goal--action pairs are then sampled around each successful pass.

\begin{algorithm}[t]
\caption{TRACE data generation}
\label{alg:datagen}
\begin{algorithmic}[1]
\Require reference $w_{\mathrm{ref}}$; targets $\mathcal{G}$; passes $P$
\State Initialize stored swings and tip paths $\mathcal{S}$ from $(g_{\mathrm{ref}},w_{\mathrm{ref}})$
\For{$P$ passes}
  \ForAll{unsolved $g\in\mathcal{G}$, nearest reference first}
    \State Choose $w_0$ from $\mathcal{S}$ with the closest tip path to $g$
    \State $w^\star\gets\textsc{CEM}(w_0; g,J)$
    \State If $M(w^\star)\le1$, add $(g,w^\star)$ to $\mathcal{S}$
  \EndFor
\EndFor
\State Refine solved targets from neighboring actions in spanning-tree order; replace only if $M\le1$.
\State \Return $\mathcal{S}$ hindsight-labeled on $M\le1$ segments
\end{algorithmic}
\end{algorithm}

\begin{figure}[t]
  \centering
    \includegraphics[width=\columnwidth]{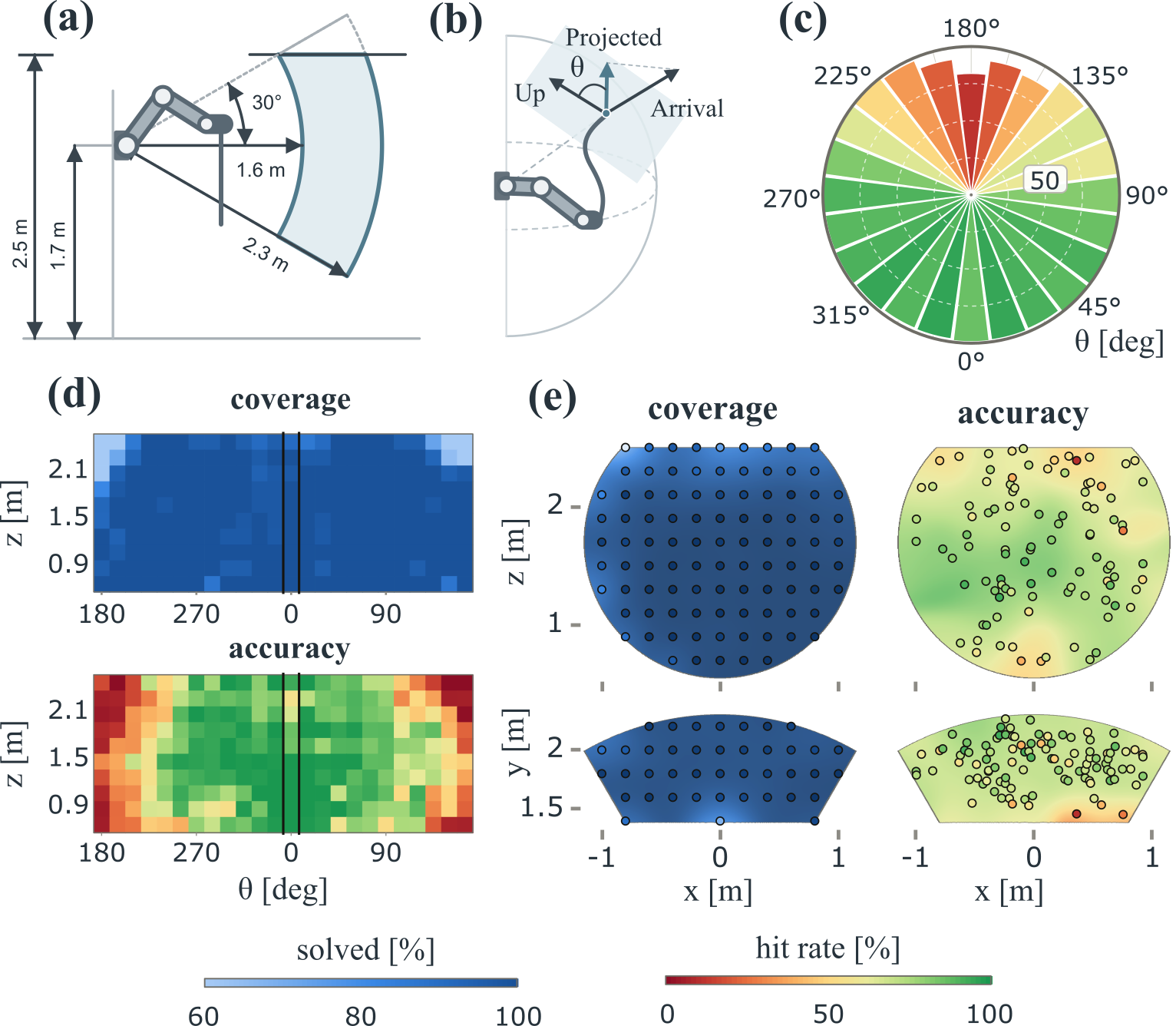}
    \caption{Experimental setup and simulation results. (a) Targets are selected within the workspace. (b) Arrival angle $\theta$, measured after projecting the rope-tip velocity onto the tangent plane of the arm-centered sphere. (c) Radar-plot radius indicates data-generation coverage for each $\theta$, and color indicates policy hit rate. (d,e) Data-generation coverage and policy accuracy, measured by solved generation targets and hit evaluation targets, respectively.}
  \label{fig:coverage}
\vspace{-0.5em}
\end{figure}



\begin{table*}[b!]
\centering
\small
\setlength{\tabcolsep}{4pt}
\begin{tabular}{llccccccccccc}
\toprule
 & & \multicolumn{5}{c}{\textbf{3D goal} (position)} & \multicolumn{6}{c}{\textbf{4D goal} (position and direction)} \\
\cmidrule(lr){3-7}\cmidrule(l){8-13}
\textbf{Rope} & \textbf{Method} & \textbf{SR@5} & \textbf{SR@10} & \textbf{Distance} & \textbf{Best} & \textbf{Worst} & \textbf{SR@5} & \textbf{SR@10} & \textbf{Distance} & \textbf{Best} & \textbf{Worst} & \textbf{Angle} \\
 & & (\%) & (\%) & (cm) & (cm) & (cm) & (\%) & (\%) & (cm) & (cm) & (cm) & ($^\circ$) \\
\midrule
\multirow{3}{*}{Rope A}    & Base           & 68 & 91 & $4.8_{\pm 4.9}$ & $1.3_{\pm 1.1}$ & $15.4_{\pm 5.9}$ & 23 & 41 & $13.4_{\pm 12.0}$ & $1.4_{\pm 0.5}$ & $38.7_{\pm 25.4}$ & $9.8_{\pm 17.3}$ \\
                           & Baseline       & 65 & 96 & $4.4_{\pm 2.8}$ & $1.3_{\pm 0.4}$ & $11.9_{\pm 2.5}$ & 31 & 45 & $11.7_{\pm 9.3}$ & $2.6_{\pm 0.4}$ & $31.0_{\pm 0.7}$ & $9.4_{\pm 13.8}$ \\
\rowcolor{gray!12}                           & \textbf{Ours}  & \textbf{89} & \textbf{99} & $\mathbf{2.8_{\pm 2.6}}$ & $\mathbf{1.0_{\pm 0.8}}$ & $\mathbf{9.8_{\pm 7.0}}$ & \textbf{61} & \textbf{75} & $\mathbf{5.2_{\pm 4.3}}$ & $\mathbf{0.7_{\pm 0.1}}$ & $\mathbf{15.2_{\pm 2.3}}$ & $\mathbf{4.5_{\pm 5.2}}$ \\
\midrule
\multirow{3}{*}{Rope B}    & Base           & 83 & 97 & $3.4_{\pm 2.8}$ & $0.9_{\pm 0.3}$ & $10.4_{\pm 1.7}$ & 33 & 55 & $9.5_{\pm 7.5}$ & $1.4_{\pm 0.7}$ & $26.1_{\pm 2.2}$ & $8.3_{\pm 11.9}$ \\
                           & Baseline       & 83 & 96 & $3.4_{\pm 2.5}$ & $\mathbf{0.5_{\pm 0.3}}$ & $11.1_{\pm 2.5}$ & 31 & 53 & $9.5_{\pm 7.3}$ & $1.0_{\pm 0.2}$ & $25.3_{\pm 1.7}$ & $8.9_{\pm 14.4}$ \\
\rowcolor{gray!12}                           & \textbf{Ours}  & \textbf{95} & \textbf{99} & $\mathbf{2.3_{\pm 1.9}}$ & $1.1_{\pm 0.7}$ & $\mathbf{8.3_{\pm 3.9}}$ & \textbf{63} & \textbf{83} & $\mathbf{5.2_{\pm 4.9}}$ & $\mathbf{1.0_{\pm 0.6}}$ & $\mathbf{20.4_{\pm 9.0}}$ & $\mathbf{5.7_{\pm 9.0}}$ \\
\midrule
\multirow{3}{*}{Rope C}    & Base           & 65 & 89 & $5.2_{\pm 4.7}$ & $\mathbf{1.4_{\pm 0.1}}$ & $19.0_{\pm 3.6}$ & 28 & 53 & $8.9_{\pm 5.8}$ & $1.6_{\pm 1.5}$ & $22.9_{\pm 0.3}$ & $6.7_{\pm 9.8}$ \\
                           & Baseline       & 73 & 91 & $4.8_{\pm 3.9}$ & $1.7_{\pm 0.6}$ & $18.6_{\pm 3.1}$ & 23 & 53 & $9.3_{\pm 5.9}$ & $\mathbf{1.4_{\pm 0.7}}$ & $22.4_{\pm 1.9}$ & $7.4_{\pm 10.5}$ \\
\rowcolor{gray!12}                           & \textbf{Ours}  & \textbf{77} & \textbf{96} & $\mathbf{4.1_{\pm 3.1}}$ & $1.5_{\pm 0.3}$ & $\mathbf{15.2_{\pm 2.9}}$ & \textbf{31} & \textbf{79} & $\mathbf{6.3_{\pm 3.3}}$ & $2.7_{\pm 0.2}$ & $\mathbf{15.7_{\pm 1.6}}$ & $\mathbf{5.0_{\pm 4.2}}$ \\
\midrule
\multirow{3}{*}{\textit{Mean}} & Base           & 72 & 92 & 4.5 & 1.2 & 14.9 & 28 & 50 & 10.6 & \textbf{1.5} & 29.2 & 8.2 \\
                           & Baseline       & 74 & 94 & 4.2 & \textbf{1.1} & 13.9 & 28 & 51 & 10.2 & 1.7 & 26.2 & 8.5 \\
\rowcolor{gray!12}                           & \textbf{Ours}  & \textbf{87} & \textbf{98} & \textbf{3.1} & 1.2 & \textbf{11.1} & \textbf{52} & \textbf{79} & \textbf{5.6} & 1.5 & \textbf{17.1} & \textbf{5.1} \\
\bottomrule
\end{tabular}
\setcounter{table}{1}
\caption{Hardware striking on three ropes, 25 targets $\times$ 3 swings per cell. Base: nominal simulator; Baseline: simulator identified by a learned network as in Wiggle\&Go~\cite{wigglego}, no correction; Ours: RECAP. SR@$d$: success rate (\%) within $d$ of the target; 4D goals also require arriving within $10^\circ$ of the commanded direction at $\ge 1$\,m/s along it. Distance and Angle (arrival-angle error): mean over all swings, with the std as a subscript. Best and Worst: the lowest and the highest per-target mean distance, with that target's std. \textit{Mean} is the mean over the three ropes. Best method per rope and goal type in bold.}
\label{tab:exp3_calib}
\end{table*}

\subsubsection{\textbf{Flow matching model}}
\label{sec:policy}

Conditional flow matching (CFM) learns to transform Gaussian noise into swing parameters~\cite{fm,icfm}. We normalize each action $w$ dimension using its training-set mean and standard deviation to get normalized action $x$. Given noise $x_0\sim\mathcal{N}(0,I)$, a normalized training action $x_1$ paired with $g$, and flow time $s\sim\mathcal{U}[0,1]$, we form $x_s=(1-s)x_0+sx_1$ and minimize
\begin{equation}
\label{eq:cfm}
\mathcal{L}(\phi)=\mathbb{E}\!\left[\left\|v_\phi(x_s,s,g)-(x_1-x_0)\right\|^2\right].
\end{equation}

The velocity network receives goal and flow-time conditioning in every residual block through adaLN-Zero~\cite{dit}. The goal's direction component is encoded as $sin(\theta)$ and $cos(\theta)$ to avoid discontinuities at angle wraparound; flow time uses a sinusoidal encoding. These encodings form a vector $u$, which controls feature scaling, shifting, and a gated residual update:
\begin{equation}
\label{eq:block}
h'\leftarrow h+\alpha_u\odot f\!\left(\mathrm{FiLM}_u(\mathrm{LN}(h))\right),
\end{equation}
where $h$ is the hidden state, $f$ is an MLP, and $\alpha_u$ is a conditioning-dependent gate. FiLM applies feature-wise scaling and shifting~\cite{film}. The modulation outputs, gates, and final output head are zero-initialized. We use 6 adaLN blocks of width $512$ ($11.3$\,M parameters), as illustrated in Fig. \ref{fig:network_architectures}.

At inference, sampling provides small variations in the predicted swing that can improve target accuracy through simulated selection. For each goal, we integrate the learned velocity field from 64 independent noise samples using 50 Euler steps, simulate these candidates, and select the action $w$ with the lowest $J$ loss. Policy generation and action correction use the same conditional flow-matching architecture, with input dimensions adapted to their conditioning variables: $g$ for policy generation and $(g,\bar{w},\Delta,\hat{\eta})$ for action correction.

\begin{table}[t]
\centering
\small
\setlength{\tabcolsep}{4pt}
\resizebox{\columnwidth}{!}{%
\begin{tabular}{@{}llccc@{}}
\toprule
\textbf{Engine} & \textbf{Envs} & \begin{tabular}[c]{@{}c@{}}\textbf{Update rate}\\(Hz)\end{tabular}
& \begin{tabular}[c]{@{}c@{}}\textbf{Step time}\\(ms)\end{tabular}
& \begin{tabular}[c]{@{}c@{}}\textbf{Sim-to-wall}\\\textbf{time ratio}\end{tabular} \\
\midrule
DeformX~\cite{deformx} (CPU, 1 thread) & 1 & 12.9 & 77.5 & 0.22 \\
\textbf{Ours} (RTX 4090) & 1 & 136 & 7.4 & 2.3 \\
\textbf{Ours} (RTX 4090) & 1024 & 104 & 9.6 & 1780 \\
\textbf{Ours} (RTX 4090) & 2048 & 77.3 & 12.9 & 2640 \\
\textbf{Ours} (RTX 4090) & 8192 & 34.5 & 29.0 & 4710 \\
\bottomrule
\end{tabular}}
\setcounter{table}{0}
\caption{Simulation throughput on the same 1\,m, 20-segment rope, on one RTX 4090. Rates are per control step of the whole batch.
}
\label{tab:throughput}
\end{table}

\begin{figure*}[th]
\centering
\includegraphics[width=1\linewidth]{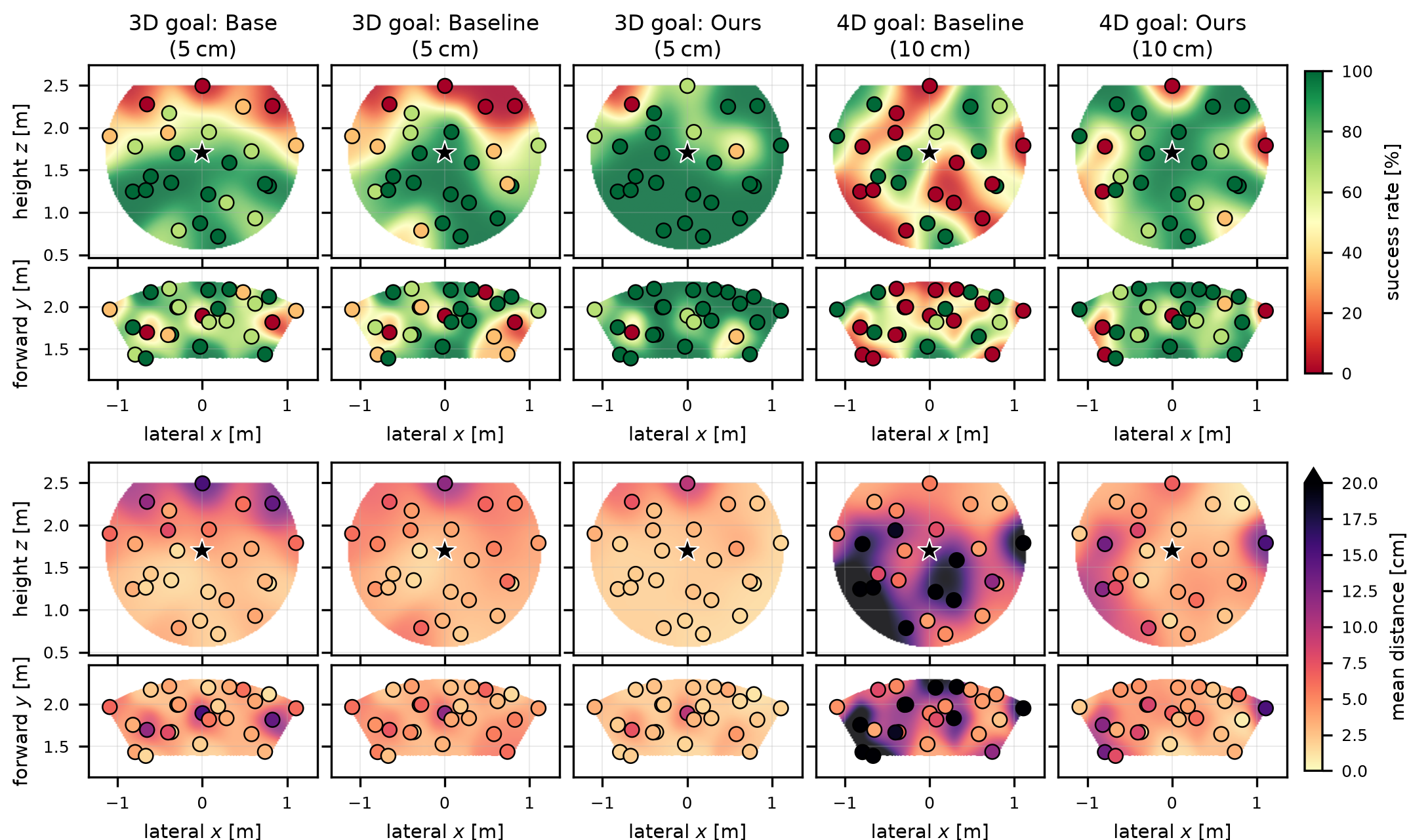}
\caption{Workspace maps on Rope~A (real world) for 3D goals with Base, Baseline and Ours, and for 4D goals with Baseline and Ours: success rate (upper two rows) and mean tip--target distance (lower two rows) of each target over its three swings. A 3D goal succeeds within $5$\,cm; a 4D goal, which also requires arriving within $10^\circ$ of the commanded direction at $\ge 1$\,m/s, is scored within $10$\,cm, as the column titles say. The distance scale saturates at $20$\,cm. Each quantity is shown in the front view (lateral $x$ vs.\ height $z$ from the wall-mounted base) and, directly below it on the same $x$ axis, in the top view (lateral $x$ vs.\ forward $y$). Dots are the evaluated targets; the shaded field is a kernel-weighted interpolation between them, and the black star marks the wall-mounted robot base.}
\label{fig:exp3_calib_map}
\vspace{-0.5em}
\end{figure*}

\subsection{RECAP: residual calibration policy}
\label{sec:calibration}

The flow-matching policy of Sec.~\ref{sec:policy} is trained on data from the
nominal simulator $\bar{\eta}$. Directly running action $\bar{w}$ from the policy generation CFM would likely fail on a real robot due to sim-to-real gaps. We designed RECAP to identify and compensate for these gaps by executing a few calibration swings for each new rope, fitting parameters to this observation $c$, and using an action correction CFM to produce corrected action $w_\eta$.


\textbf{System identification.} The robot executes 8 fixed calibration
swings, while motion capture records the rope-tip trajectories $c$.
Through experiments, we identified 11 parameters $\eta$ that we can tune to approximate the real setup, which helps us build an approximate simulator.
We then use CEM to search for the parameters $\hat\eta$ that best reproduce these
trajectories in simulation, minimizing the difference between $c$ and $c_{\hat\eta}$, as shown in Fig. \ref{fig:network_architectures}(b). 


\textbf{Delta policy.} For a goal $g$, policy generation CFM produces the nominal action $\bar{w}$. The delta policy is a conditional flow-matching model with the same architecture that learns to generate real action $w_\eta$ conditioned on the real goal and system from $\bar{w}$.
\begin{equation}
\begin{aligned}
w_\eta \sim p_\psi\!\left(w_{\hat{\eta}} \mid g,\, \bar{w},\, \Delta_{\hat{\eta}}(\bar{w}),\, \hat{\eta}\right)
\end{aligned}
\end{equation}
where $\Delta_{\hat{\eta}}(\bar{w})$ is the error to the goal at the pass of $\bar{w}$ in the fitted simulator. The delta policy is trained on simulated ropes and rigs with perturbed parameters. Its labels are corrections that CEM finds to improve the pass.

During inference, we generate eight correction actions, replay those, and the nominal action in the fitted simulator, and execute the candidate with the lowest loss $J$. Thus, each new rope only costs 8 real swings, and every goal afterwards takes a single strike, with no measurement during execution.

\subsection{DeformX 2.0: the GPU co-simulation engine}
\label{sec:engine}

DeformX~\cite{deformx} couples a Cosserat-rod solver to Isaac Sim for accurate DLO dynamics but is limited by CPU throughput. To support TRACE search, correction-policy training, and candidate verification, we adopt a GPU Cosserat-rod formulation~\cite{cosserat} that separates position and rotation updates and solves orientation in closed form, enabling stable integration at large time steps and  efficient batched simulation.

We use 100 solver substeps per 60\,Hz control step with 16 constraint iterations, simulating thousands of rope worlds in parallel with the robot arm in Isaac Sim~\cite{isaac_Sim}.

To capture aerodynamic effects at swing speeds, we add quadratic cross-flow drag~\cite{air_drag,flycast}, damping each segment's velocity normal to the local rope tangent at every substep.

DeformX~2.0 generates over 1 million trajectories per hour per GPU. With $8192$ parallel environments, its throughput exceeds DeformX by more than $20{,}000\times$ (Table~\ref{tab:throughput}).

\section{Experiment}

The experiments answer three questions: how fast the simulation is, how accurately the policy strikes a commanded position and arrival direction in simulation and on hardware, and how much the data generation and the rope calibration each contribute. All arms share the setup and protocol below.

\subsection{Experiment Setup}

We evaluate the policy in simulation and on a wall-mounted 6-DoF UR5e holding a $0.65$\,m rigid stick with a $1.0$\,m rope. The workspace is a $30^\circ$ cone about the base axis, with radii $[1.6,2.3]$\,m and a maximum height of $2.5$m, as shown in Fig. \ref{fig:coverage} (a). Joint velocities are limited to $2$\,rad/s.

The policy uses a 21-parameter action: four evenly timed waypoints for five joints and the swing duration. The final wrist joint is excluded because the stick lies along its rotation axis and have negligible effect on the swing.


Hardware evaluation uses 25 targets with three swings each from separate policy samples. Simulation uses 5700 uniformly sampled workspace targets shared across policies to measure coverage and error.

\subsection{Hit Target Experiment}
\label{sec:exp1}

This experiment evaluates striking on hardware with three ropes that differ in
material, diameter and linear density, as shown in Fig.~\ref{fig:figure1}. Each rope is tested on the same 25
targets with two goal types. A 3D goal specifies only the target position. A 4D
goal also specifies the arrival direction, drawn at random for each target.

We compare three methods, all of which start from samples of the same policy. Base executes the best candidate in the nominal simulator. Baseline follows Wiggle\&Go~\cite{wigglego}: a simulation-trained network estimates rope and rig parameters, and the best policy candidate in the identified simulator is executed. Ours is RECAP (Sec.~\ref{sec:calibration}), fitting the simulator from eight calibration swings recorded once per rope.

A swing succeeds at radius $d$ when the rope tip passes within $d$ of the
target, and we report $d=5$ and $10$\,cm. A 4D swing must also arrive within
$10^\circ$ of the commanded direction, at no less than $1$\,m/s along it.

Table~\ref{tab:exp3_calib} reports the results. RECAP is the strongest method
on every rope and both goal types: averaged over the three ropes it raises
SR@5\,cm from $72\%$ to $87\%$ on 3D goals and from $28\%$ to $52\%$ on 4D
goals, against $74\%$ and $28\%$ for Baseline, and it halves the mean 4D miss
($10.6 \to 5.6$\,cm). The advantage grows with the difficulty of the goal: it
is widest on 4D goals, and even on Rope~C, RECAP brings most 4D misses inside
$10$\,cm ($53\%$ to $79\%$).

Fig.~\ref{fig:exp3_calib_map} shows how success rate and miss distance vary across the Rope~A workspace. Both degrade toward the upper edge for Base and Baseline, while RECAP largely removes this degradation, leaving only the top corner weak. Overall, a one-time eight-swing calibration transfers one simulation-trained policy to three different ropes.

\begin{table}[t]
\centering\small
\setlength{\tabcolsep}{2pt}
\begin{tabular}{lccccc}
    \toprule
    \textbf{Method} & \textbf{Params} & \textbf{One-shot} &
    \textbf{Best-of-64} & \textbf{Miss} & \textbf{Angle} \\
    & (M) & (\%) & (\%) & (cm) & ($^\circ$) \\
    \midrule
    \textbf{Ours}     & 11.32 & \textbf{73.5} & \textbf{92.1} & \textbf{3.6} & \textbf{2.4} \\
    Naive FM          & 11.35 & 67.9 & 88.9 & 5.3  & 3.0 \\
    Regression        & 11.31 & 21.4 & --   & 24.0 & 12.4 \\
    \midrule
    Nearest-neighbor & ---   & 33.6 & --   & 9.2  & 6.4 \\
    \bottomrule
\end{tabular}
\setcounter{table}{2}
\caption{\textbf{Learning-method ablation}: We compare our model, naive flow matching, and our network with a regression head at matched capacity, alongside nearest-neighbor lookup in the training data. Miss and angle are means over the one-shot sample on $n=5700$ held-out goals. Regression falls below the lookup because it averages incompatible motion patterns into invalid swings.}
\label{tab:fm_ablation}
\end{table}

\section{ABLATION STUDY}

\subsection{Learning distinct motion patterns}

We evaluate how the policy architecture represents distinct motion patterns for the same goal. We compare our model with a regression baseline
that replaces our flow-matching head with a regression head, a naively conditioned flow-matching model that receives the goal only at its input, and a nearest-neighbor baseline that retrieves the swing associated with the closest training goal. Learned models use matched capacity and
training protocols (Table~\ref{tab:fm_ablation}).

\begin{figure}[h]
  \centering
  \includegraphics[width=0.95\linewidth]{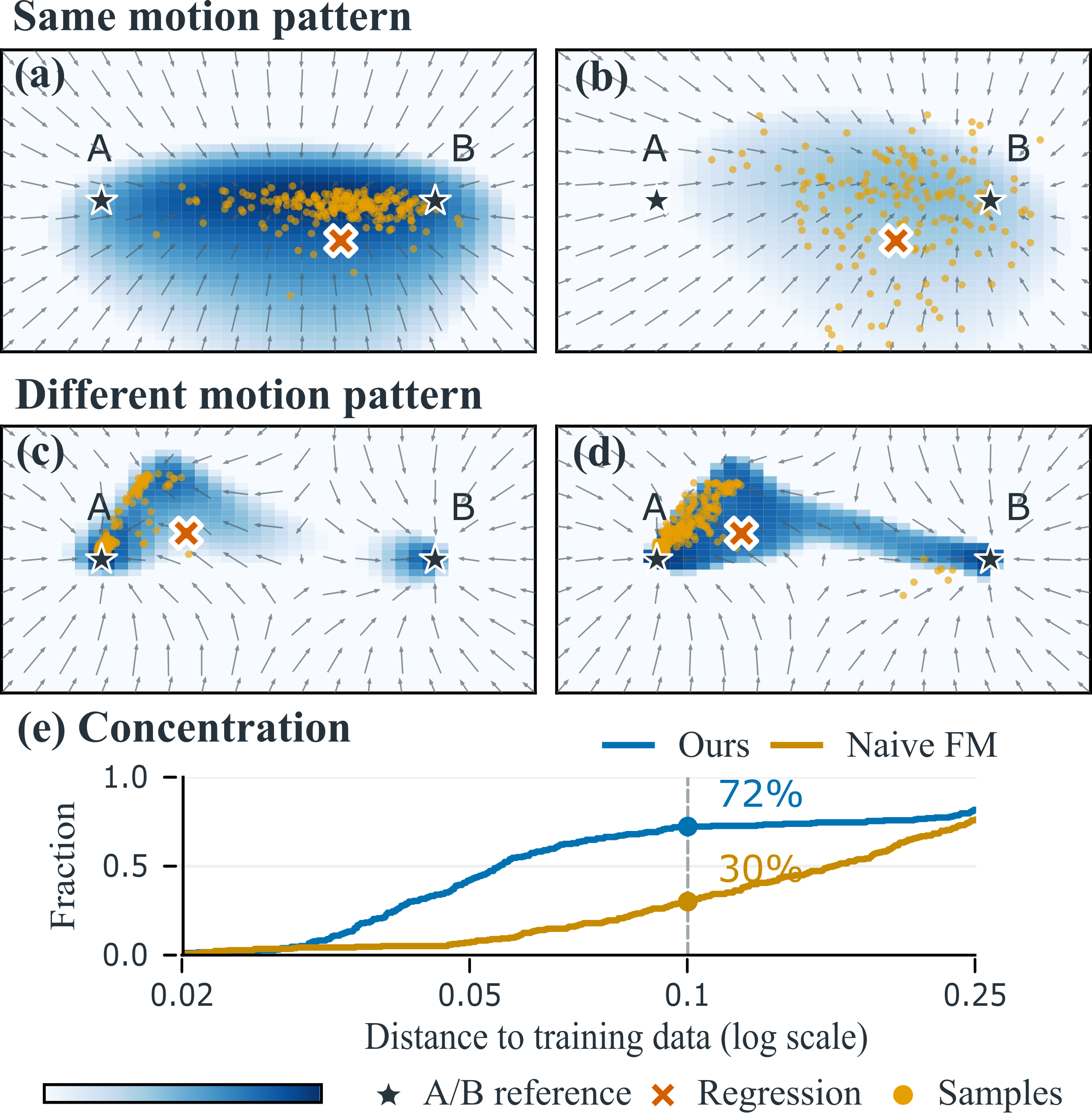}
  \caption{Learning multiple swings for one goal. We compare goal conditioning in every block (ours, left) with conditioning only at the input (naive FM, right). Stars mark two training swings \(w_A,w_B\) associated with goal \(g\); dots show generated samples. Shading shows a 2D slice of the conditional log-density \(\log p(w\mid g)\), and arrows show the learned flow field at \(s=0.9\). (a,b) For swings from the same motion pattern, our model produces more concentrated samples. (c,d) For swings from different patterns, our model produces more clearly separated density peaks. Regression (\(\times\)) predicts a single action that can average incompatible swings. (e) For the pair in (c,d), curves show the fraction of samples within a given distance of the nearer training swing, normalized by the pair’s separation in normalized action space.}
  \label{fig:fm_abl}
\end{figure}

Fig.~\ref{fig:fm_abl} shows that our model keeps samples compact within each motion pattern while separating distinct patterns. Naive conditioning produces more dispersed samples, while regression can place its prediction between incompatible swings. This distinction matters because an average of successful swings may not succeed.

\subsection{Data generation for policy learning}
\label{sec:exp_datagen}

We evaluate the data-generation strategy in terms of workspace coverage,
strike accuracy, and trajectory quality. We compare our full TRACE method with CEM
initialized from an example swing and our method without an example. We assess
both the generated trajectories and the policies trained on the resulting datasets (Table~\ref{tab:exp2_datagen}).

Our TRACE method achieves broader workspace coverage and higher policy
accuracy after candidate selection. The example's contribution is reflected
in lower curvature loss and stronger performance after candidate selection.
The example-free variant achieves slightly higher one-shot accuracy,
highlighting that trajectory quality and the accuracy of an individual
policy sample are distinct properties.

\begin{table}[t]
\centering
\footnotesize
\setlength{\tabcolsep}{2pt}
\begin{tabular}{@{}lrrrr@{}}
\toprule
\textbf{Generator} & \shortstack{\textbf{Coverage}\\(\%) $\uparrow$} & \shortstack{\textbf{One-shot}\\(\%) $\uparrow$} & \shortstack{\textbf{Best-of-64}\\(\%) $\uparrow$} & \shortstack{\textbf{Mean}\\$L_{\mathrm{curv}}$ $\downarrow$} \\
\midrule
CEM w/ example & 49.5 & 58.5 & 68.5 & 0.40 \\
TRACE w/o example & 96.8 & \textbf{75.5} & 83.9 & 0.51 \\
\textbf{TRACE} & \textbf{98.3} & 73.5 & \textbf{92.1} & \textbf{0.33} \\
\bottomrule
\end{tabular}
\caption{Data-generation ablation evaluating workspace coverage, policy accuracy on 5700 random test goals, and trajectory quality measured by mean curvature loss. One-shot evaluates one policy sample; best-of-64 selects the candidate with the lowest TRACE loss. The example seeds higher-quality motion patterns, while TRACE improves coverage and accuracy of the dataset.}
\label{tab:exp2_datagen}
\vspace{-1em}
\end{table}

\section{Conclusion and Discussion}

This work extends rope-tip targeting to goals that specify both position and arrival direction (4D goals) with a zero-shot swing. Our framework learns striking motions in simulation and uses a few calibration swings to adapt them to a real system. The policy achieves $92.1\%$ success within 5 cm and 10° in simulation, while on hardware, across three ropes, calibration raises success within 5\,cm to $87\%$ for position goals and $52\%$ for position-and-direction goals.

Our experience suggests a broader insight for deformable object manipulation. 
We view the dynamics of a deformable object as a complex nonlinear mapping from robot action space to task space outcomes. Manipulation requires solving the inverse map: finding actions that produce a desired outcome. Our observations suggest that this mapping contains locally predictable regions, where small action space changes lead to gradual changes in target space. We observe this structure within motion patterns, while transitions between patterns can be highly sensitive and chaotic. Starting from a swing verified on hardware and extending the search locally produces consistent training data that makes inverse learning easier. These patterns can also benefit sim-to-real transfer by limiting how execution errors and variations in physical properties affect the outcome. Selecting such regions can simplify learning and deployment and may help explain why our policy, trained in simulation, achieves useful hardware performance even before calibration. We believe that identifying similarly robust regions could also benefit other deformable object manipulation tasks.


Improving simulation can reduce reliance on online correction in deformable object manipulation. While feedback compensates for execution errors, a sufficiently accurate simulator can generate better motions in advance. In our framework, its practical value lies in capturing the dynamics well enough to produce effective motions and reliably rank candidates. Improving simulation provides a complementary path to reliable manipulation alongside feedback control.

Several limitations remain. First, our framework targets single-swing rope-tip striking and does not yet address more general forms of dynamic deformable-object manipulation involving sustained contact, obstacle interaction, or multi-stage behavior. Second, execution remains open-loop after a one-time calibration, so disturbances or system changes that occur during a swing cannot be corrected online. Finally, although our calibration reduces the need for an exact physical model, the method still relies on simulation to generate and rank candidate motions, and its robustness may degrade when the real system lies substantially outside the range of dynamics represented during training.

Extending the framework toward closed-loop feedback and broader classes of dynamic DLO interactions is an important direction for future work.


\section*{Acknowledgment}
The authors used ChatGPT and Claude to assist with language editing and improving the clarity of the manuscript. All technical content, analyses, figures, results, and conclusions were produced and verified by the authors.

\bibliography{refs}
\bibliographystyle{IEEEtran}


\end{document}